\documentclass[10pt,twocolumn,letterpaper]{article}

\usepackage{cvpr}
\usepackage{times}
\usepackage{epsfig}
\usepackage{graphicx}
\usepackage{amsmath}
\usepackage{amssymb}
\usepackage{lineno}
\usepackage{multirow}
\usepackage{bbm}
\usepackage{caption}
\usepackage{subcaption}
\DeclareMathOperator*{\argmin}{argmin}

\usepackage[pagebackref=true,breaklinks=true,letterpaper=true,colorlinks,bookmarks=false]{hyperref}

\cvprfinalcopy 

\def\cvprPaperID{255} 

\ifcvprfinal\fi
\begin{document}

\title{Visual Saliency Based on Multiscale Deep Features}

\author{Guanbin Li \hspace{1.0in} Yizhou Yu\\
Department of Computer Science, The University of Hong Kong\\
{\tt\small https://sites.google.com/site/ligb86/mdfsaliency/ }
}

\maketitle

\begin{abstract}
Visual saliency is a fundamental problem in both cognitive and computational sciences, including computer vision. In this paper, we discover that a high-quality visual saliency model can be learned from multiscale features extracted using deep convolutional neural networks (CNNs), which have had many successes in visual recognition tasks. For learning such saliency models, we introduce a neural network architecture, which has fully connected layers on top of CNNs responsible for feature extraction at three different scales. We then propose a refinement method to enhance the spatial coherence of our saliency results. Finally, aggregating multiple saliency maps computed for different levels of image segmentation can further boost the performance, yielding saliency maps better than those generated from a single segmentation. To promote further research and evaluation of visual saliency models, we also construct a new large database of 4447 challenging images and their pixelwise saliency annotations. Experimental results demonstrate that our proposed method is capable of achieving state-of-the-art performance on all public benchmarks, improving the F-Measure by 5.0\% and 13.2\% respectively on the MSRA-B dataset and our new dataset (HKU-IS), and lowering the mean absolute error by 5.7\% and 35.1\% respectively on these two datasets.
\end{abstract}

\section{Introduction}\label{sec:intro}
Visual saliency attempts to determine the amount of attention steered towards various regions in an image by the human visual and cognitive systems~\cite{BI13}. It is thus a fundamental problem in psychology, neural science, and computer vision. Computer vision researchers focus on developing computational models for either simulating the human visual attention process or predicting visual saliency results. 
Visual saliency has been incorporated in a variety of computer vision and image processing tasks to improve their performance. Such tasks include image cropping~\cite{Autocollage}, retargeting~\cite{SeamCarving}, and summarization~\cite{SCSI08}.
Recently, visual saliency has also been increasingly used by visual recognition tasks~\cite{RWKP04}, such as image classification~\cite{WYW13} and person re-identification~\cite{ZOW13}.

Human visual and cognitive systems involved in the visual attention process are composed of layers of interconnected neurons. For example, the human visual system has layers of simple and complex cells whose activations are determined by the magnitude of input signals falling into their receptive fields. Since deep artificial neural networks were originally inspired by biological neural networks,
it is thus a natural choice to build a computational model of visual saliency using deep artificial neural networks. Specifically, recently popular convolutional neural networks (CNN) are particularly well suited for this task because convolutional layers in a CNN resemble simple and complex cells in the human visual system \cite{fukushima1980neocognitron} while fully connected layers in a CNN resemble higher-level inference and decision making in the human cognitive system.

In this paper, we develop a new computational model for visual saliency using multiscale deep features computed by convolutional neural networks. Deep neural networks, such as CNNs, have recently achieved many successes in visual recognition tasks~\cite{krizhevsky2012imagenet,FCNL13,girshick2013rich,hariharan2014simultaneous}.
Such deep networks are capable of extracting feature hierarchies from raw pixels automatically. Further, features extracted using such networks are highly versatile and often more effective than traditional handcrafted features. Inspired by this, we perform feature extraction using a CNN originally trained over the ImageNet dataset~\cite{deng2009imagenet}. Since ImageNet contains images of a large number of object categories, our features contain rich semantic information, which is useful for visual saliency because humans pay varying degrees of attention to objects from different semantic categories.
For example, viewers of an image likely pay more attention to objects like cars than the sky or grass. In the rest of this paper, we call such features {\em CNN features}.

By definition, saliency is resulted from visual contrast as it intuitively characterizes certain parts of an image that appear to stand out relative to their neighboring regions or the rest of the image.
Thus, to compute the saliency of an image region, our model should be able to evaluate the contrast between the considered region and its surrounding area as well as the rest of the image. Therefore, we extract multiscale CNN features for every image region from three nested and increasingly larger rectangular windows, which respectively encloses the considered region, its immediate neighboring regions, and the entire image.

On top of the multiscale CNN features, our method further trains fully connected neural network layers. Concatenated multiscale CNN features are fed into these layers trained using a collection of labeled saliency maps. Thus, these fully connected layers play the role of a regressor that is capable of inferring the saliency score of every image region from the multiscale CNN features extracted from nested windows surrounding the image region. It is well known that deep neural networks with at least one fully connected layers can be trained to achieve a very high level of regression accuracy.

We have extensively evaluated our CNN-based visual saliency model over existing datasets, and meanwhile noticed a lack of large and challenging datasets for training and testing saliency models. At present, the only large dataset that can be used for training a deep neural network based model was derived from the MSRA-B dataset~\cite{liu2011learning}. This dataset has become less challenging over the years because images there typically include a single salient object located away from the image boundary. To facilitate research and evaluation of advanced saliency models, we have created a large dataset where an image likely contains multiple salient objects, which have a more general spatial distribution in the image.
Our proposed saliency model has significantly outperformed all existing saliency models over this new dataset as well as all existing datasets.

In summary, this paper has the following contributions:
\begin{itemize}
\item A new visual saliency model is proposed to incorporate multiscale CNN features extracted from nested windows with a deep neural network with multiple fully connected layers.
    The deep neural network for saliency estimation is trained using regions from a set of labeled saliency maps.
\item A complete saliency framework is developed by further integrating our CNN-based saliency model with a spatial coherence model and multi-level image segmentations.
\item A new challenging dataset, HKU-IS, is created for saliency model research and evaluation.
This dataset is publicly available.
Our proposed saliency model has been successfully validated on this new dataset as well as on all existing datasets.
\end{itemize}

\subsection{Related Work}\label{sec:related}
Visual saliency computation can be categorized into bottom-up and top-down methods or a hybrid of the two. Bottom-up models are primarily based on a center-surround scheme, computing a master saliency map by a linear or non-linear combination of low-level visual attributes such as color, intensity, texture and orientation~\cite{itti1998model,hou2007saliency,achanta2009frequency,ChengPAMI,liu2011learning}. Top-down methods generally require the incorporation of high-level knowledge, such as objectness and face detector in the computation process~\cite{jia2013category,chang2011fusing,goferman2012context,shen2012unified,liu2014adaptive}.

Recently, much effort has been made to design discriminative features and saliency priors.
Most methods essentially follow the region contrast framework, aiming to design features that better characterize the distinctiveness of an image region with respect to its surrounding area. In \cite{liu2011learning}, three novel features are integrated with a conditional random field. A model based on low-rank matrix recovery is presented in \cite{shen2012unified} to integrate low-level visual features with higher-level priors.

Saliency priors, such as the center prior~\cite{liu2011learning,wei2012geodesic,judd2009learning} and the boundary prior~\cite{jiang2013salient,zhu2014saliency}, are widely used to heuristically combine low-level cues and improve saliency estimation.
These saliency priors are either directly combined with other saliency cues as weights~\cite{ChengPAMI,cheng2013efficient,jia2013category} or used as features in learning based algorithms~\cite{jiang2013salient,judd2009learning,liu2014adaptive}. While these empirical priors can improve saliency results for many images, they can fail when a salient object is off-center or significantly overlaps with the image boundary. Note that object location cues and boundary-based background modeling are not neglected in our framework, but have been implicitly incorporated into our model through multiscale CNN feature extraction and neural network training.

Convolutional neural networks have recently achieved many successes in visual recognition tasks, including image classification~\cite{krizhevsky2012imagenet}, object detection~\cite{girshick2013rich}, and scene parsing~\cite{FCNL13}. Donahue et al.\cite{donahue2013decaf} pointed out that features extracted from Krizhevsky's CNN trained on the ImageNet dataset~\cite{deng2009imagenet} can be repurposed to generic tasks. Razavian et al.\cite{razavian2014cnn} extended their results and concluded that deep learning with CNNs can be a strong candidate for any visual recognition task. Nevertheless, CNN features have not yet been explored in visual saliency research primarily because saliency cannot be solved using the same framework considered in \cite{donahue2013decaf,razavian2014cnn}. It is the contrast against the surrounding area rather than the content inside an image region that should be learned for saliency prediction. This paper proposes a simple but very effective neural network architecture to make deep CNN features applicable to saliency modeling and salient object detection.

\begin{figure}[ht]
\begin{center}
   \includegraphics[width=0.5\textwidth]{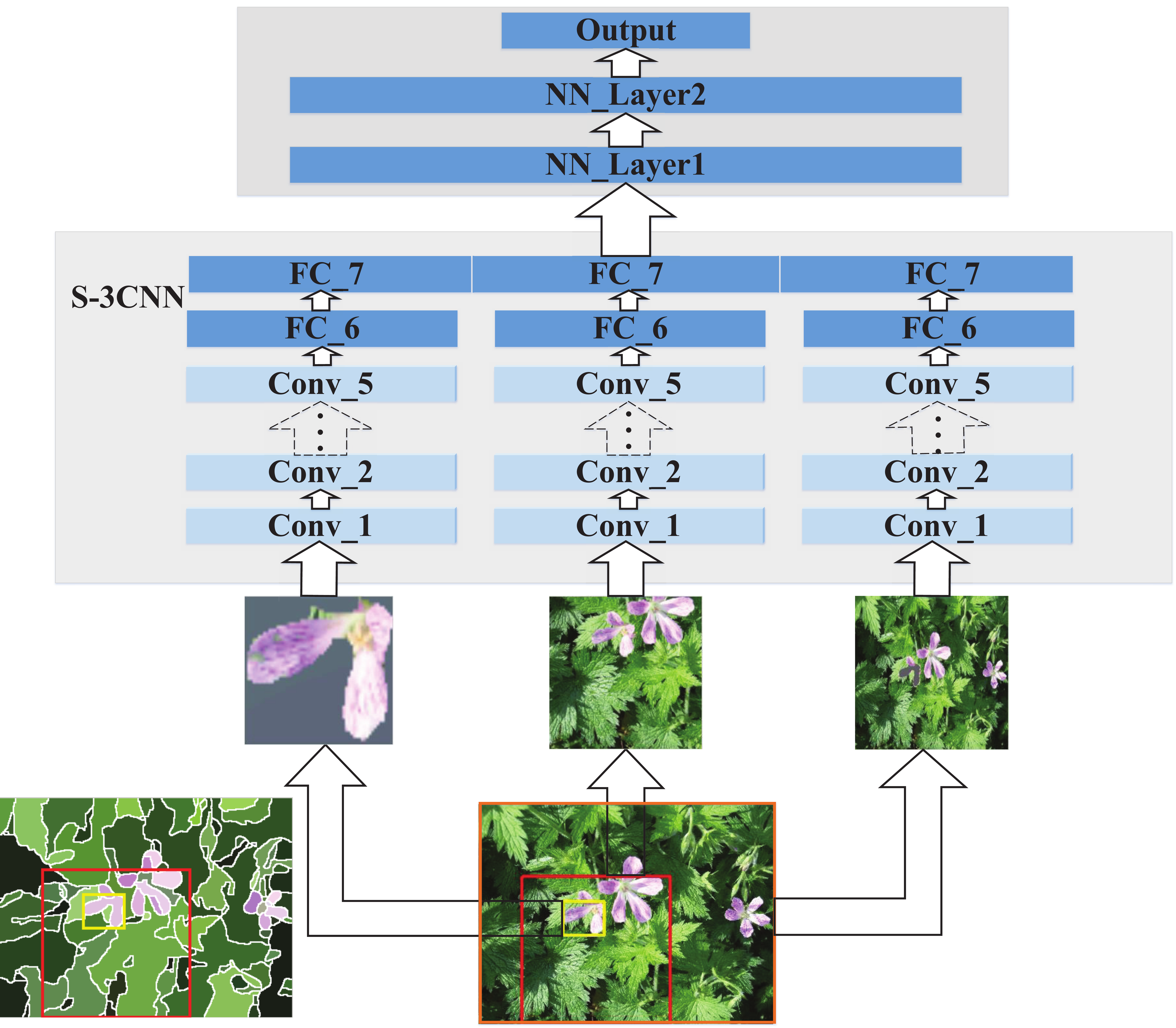}
\end{center}
   \caption{The architecture of our deep feature based visual saliency model.}
\label{fig:arch}
\end{figure}

\section{Saliency Inference with Deep Features}\label{sec:deep}
As shown in Fig. \ref{fig:arch}, the architecture of our deep feature based model for visual saliency consists of one output layer and two fully connected hidden layers on top of three deep convolutional neural networks. Our saliency model requires an input image to be decomposed into a set of nonoverlapping regions, each of which has almost uniform saliency values internally. The three deep CNNs are responsible for multiscale feature extraction. For each image region, they perform automatic feature extraction from three nested and increasingly larger rectangular windows, which are respectively the bounding box of the considered region, the bounding box of its immediate neighboring regions, and the entire image. The features extracted from the three CNNs are fed into the two fully connected layers, each of which has 300 neurons. The output of the second fully-connected layer is fed into the output layer, which performs two-way softmax that produces a distribution over binary saliency labels. When generating a saliency map for an input image, we run our trained saliency model repeatedly over every region of the image to produce a single saliency score for that region. This saliency score is further transferred to all pixels within that region.


\subsection{Multiscale Feature Extraction}\label{sec:feature}
We extract multiscale features for each image region with a deep convolutional neural network originally trained over the ImageNet dataset~\cite{deng2009imagenet} using Caffe~\cite{jia2014caffe}, an open source framework for CNN training and testing. The architecture of this CNN has eight layers including five convolutional layers and three fully-connected layers. Features are extracted from the output of the second last fully connected layer, which has 4096 neurons. Although this CNN was originally trained on a dataset for visual recognition, automatically extracted CNN features turn out to be highly versatile and can be more effective than traditional handcrafted features on other visual computing tasks.

Since an image region may have an irregular shape while CNN features have to be extracted from a rectangular region, to make the CNN features only relevant to the pixels inside the region, as in \cite{girshick2013rich}, we define the rectangular region for CNN feature extraction to be the bounding box of the image region and fill the pixels outside the region but still inside its bounding box with the mean pixel values at the same locations across all ImageNet training images. These pixel values become zero after mean subtraction and do not have any impact on subsequent results.
We warp the region in the bounding box to a square with 227x227 pixels to make it compatible with the deep CNN trained for ImageNet. The warped RGB image region is then fed to the deep CNN and a 4096-dimensional feature vector is obtained by forward propagating a mean-subtracted input image region through all the convolutional layers and fully connected layers. We name this vector {\em feature A}.


Feature A itself does not include any information around the considered image region, thus is not able to tell whether the region is salient or not with respect to its neighborhood as well as the rest of the image. To include features from an area surrounding the considered region for understanding the amount of contrast in its neighborhood, we extract a second feature vector from a rectangular neighborhood, which is the bounding box of the considered region and its immediate neighboring regions. All the pixel values in this bounding box remain intact. Again, this rectangular neighborhood is fed to the deep CNN after being warped. We call the resulting vector from the CNN {\em feature B}.

As we know, a very important cue in saliency computation is the degree of (color and content) uniqueness of a region with respect to the rest of the image. The position of an image region in the entire image is another crucial cue. To meet these demands, we use the deep CNN to extract {\em feature C} from the entire rectangular image, where the considered region is masked with mean pixel values for indicating the position of the region. These three feature vectors obtained at different scales together define the features we adopt for saliency model training and testing. Since our final feature vector is the concatenation of three CNN feature vectors, we call it S-$3$CNN.

\subsection{Neural Network Training}\label{sec:train}
On top of the multiscale CNN features, we train a neural network with one output layer and two fully connected hidden layers. This network plays the role of a regressor that infers the saliency score of every image region from the multiscale CNN features extracted for the image region. It is well known that neural networks with fully connected hidden layers can be trained to reach a very high level of regression accuracy.


Concatenated multiscale CNN features are fed into this network, which is trained using a collection of training images and their labeled saliency maps, that have pixelwise binary saliency scores. Before training, every training image is first decomposed into a set of regions. The saliency label of every image region is further estimated using pixelwise saliency labels. During the training stage, only those regions with 70\% or more pixels with the same saliency label are chosen as training samples, and their saliency labels are set to either 1 or 0 respectively.
During training, the output layer and the fully connected hidden layers together minimize the least-squares prediction errors accumulated over all regions from all training images.

Note that the output of the penultimate layer of our neural network is indeed a fine-tuned feature vector for saliency detection. Traditional regression techniques, such as support vector regression and random forests, can be further trained on this feature vector to generate a saliency score for every image region. In our experiments, we found that this feature vector is very discriminative and the simple logistic regression embedded in the final layer of our architecture is strong enough to generate state-of-the-art performance on all visual saliency datasets.

\section{The Complete Algorithm}\label{sec:algo}
\subsection{Multi-Level Region Decomposition}\label{sec:region}
A variety of methods can be applied to decompose an image into nonoverlapping regions. Examples include grids, region growing, and pixel clustering. Hierarchical image segmentation can generate regions at multiple scales to support the intuition that a semantic object at a coarser scale may be composed of multiple parts at a finer scale. To enable a fair comparison with previous work on saliency estimation, we follow the multi-level region decomposition pipeline in~\cite{jiang2013salient}.
Specifically, for an image \emph{I}, \emph{M} levels of image segmentations, $S = \{S_1, S_2, ..., S_M\} (|S_i|=N_i)$, are constructed from the finest to the coarsest scale. The regions at any level form a nonoverlapping decomposition. The hierarchical region merge algorithm in \cite{arbelaez2011contour} is applied to build a segmentation tree for the image.
The initial set of regions are called superpixels. They are generated using the graph-based segmentation algorithm in \cite{felzenszwalb2004efficient}. Region merge is prioritized by the edge strength at the boundary pixels shared by two adjacent regions. Regions with lower edge strength between them are merged earlier. The edge strength at a pixel is determined by a real-valued ultrametric contour map (UCM). In our experiments, we normalize the value of UCM into $[0,1]$ and generate 15 levels of segmentations with different edge strength thresholds. The edge strength threshold for level $i$ is adjusted such that the number of regions reaches a predefined target. The target number of regions at the finest and coarsest levels are set to 300 and 20 respectively, and the number of regions at intermediate levels follows a geometric series.

\subsection{Spatial Coherence}\label{sec:coherence}
Given a region decomposition of an image, we can generate an initial saliency map with the neural network model presented in the previous section. However, due to the fact that image segmentation is imperfect and our model assigns saliency scores to individual regions, noisy scores inevitably appear in the resulting saliency map. To enhance spatial coherence, a superpixel based saliency refinement method is used. The saliency score of a superpixel is set to the mean saliency score over all pixels in the superpixel. The refined saliency map is obtained by minimizing the following cost function, which can be reduced to solving a linear system.
\begin{equation}
\begin{split}\label{eq:coherence}
\sum_{i} \left( \mathrm{a}^R_{i} - \mathrm{a}^I_i \right)^2
&+ \sum_{i,j}w_{ij} \left(\mathrm{a}^R_{i} - \mathrm{a}^R_{j} \right)^2,
\end{split}
\end{equation}
where $\mathrm{a}^I_i$ is the initial saliency score at superpixel $i$, $\mathrm{a}^R_i$ is the refined saliency score at the same superpixel. The first term in (\ref{eq:coherence}) encourages similarity between the refined saliency map and the initial saliency map, while the second term is an all-pair spatial coherence term that favors consistent saliency scores across different superpixels if there do not exist strong edges separating them. $w_{ij}$ is the spatial coherence weight between any pair of superpixels $P_i$ and $P_j$.

To define pairwise weights $w_{ij}$, we construct an undirected weighted graph on the set of superpixels. There is an edge in the graph between any pair of adjacent superpixels $(P_i, P_j)$, and the distance between them is defined as follows,
\begin{equation}
 d(P_i, P_j) = \frac{\sum_{p\in \left( \Omega_{P_i} \bigcap P_j \bigcup P_i \bigcap \Omega_{P_j} \right)} ES(p)}{|\Omega_{P_i} \bigcap P_j \bigcup P_i \bigcap \Omega_{P_j}|},
\end{equation}
where $ES(p)$ is the edge strength at pixel $p$ and $\Omega_P$ represents the set of pixels on the outside boundary of superpixel $P$. We again make use of the UCM proposed in \cite{arbelaez2011contour} to define edge strength here. The distance between any pair of non-adjacent superpixels is defined as the shortest path distance in the graph.
The spatial coherence weight $w_{ij}$ is thus defined as
$w_{ij} = \exp \left( -\frac{d^{2}(P_i, P_j)}{2\sigma^{2}} \right)$,
where $\sigma$ is set to the standard deviation of pairwise distances in our experiments. This weight is large when two superpixels are located in the same homogeneous region and small when they are separated by strong edges.

\subsection{Saliency Map Fusion}\label{sec:fusion}
We apply both our neural network model and spatial coherence refinement to each of the $M$ levels of segmentation. As a result, we obtain $M$ refined saliency maps, $\{A^{(1)}, A^{(2)}, ..., A^{(M)}\}$, interpreting salient parts of the input image at various granularity. We aim to further fuse them together to obtain a final aggregated saliency map.
To this end, we take a simple approach by assuming the final saliency map is a linear combination of the maps at individual segmentation levels, and learn the weights in the linear combination by running a least-squares estimator over a validation dataset, indexed with $I_v$. Thus, our aggregated saliency map $A$ is formulated as follows,
{\small
\begin{equation}
\begin{split}
&A = \sum_{k = 1}^{M} \alpha_k A^{(k)} \\
\text{s.t. }  \{\alpha_k\}_{k=1}^{M} &= \argmin_{\alpha_1, \alpha_2, ..., \alpha_M} \sum_{i\in I_v}\left\|A_i - \sum_{k} \alpha_k A_i^{(k)}\right\|^2_F.
\end{split}
\end{equation}
}

Note that there are many options for saliency fusion. For example, a conditional random field (CRF) framework has been adopted in \cite{mai2013saliency} to aggregate multiple saliency maps from different methods. 
Nevertheless, we have found that, in our context, a linear combination of all saliency maps can already serve our purposes well and is capable of producing aggregated maps with a quality comparable to those obtained from more complicated techniques.

\section{A New Dataset}\label{sec:data}
At present, the pixelwise ground truth annotation~\cite{jiang2013salient} of the MSRA-B dataset~\cite{liu2011learning} is the only large dataset that is suitable for training a deep neural network. Nevertheless, this benchmark becomes less challenging once a center prior and a boundary prior~\cite{jiang2013salient,zhu2014saliency} have been imposed since most images in the dataset contain only one connected salient region and 98\% of the pixels in the border area belongs to the background~\cite{jiang2013salient}.

We have constructed a more challenging dataset to facilitate the research and evaluation of visual saliency models. To build the dataset, we initially collected 7320 images. These images were chosen by following at least one of the following criteria:
\begin{enumerate}
\item there are multiple disconnected salient objects;\vspace{-1mm}
\item at least one of the salient objects touches the image boundary;\vspace{-1mm}
\item the color contrast (the minimum Chi-square distance between the color histograms of any salient object and its surrounding regions) is less than 0.7.
\end{enumerate}
To reduce label inconsistency, we asked three people to annotate salient objects in all 7320 images individually using a custom designed interactive segmentation tool. On average, each person takes 1-2 minutes to annotate one image. The annotation stage spanned over three months.

Let $A^p = \{a_x^{(p)}\}$ be the binary saliency mask labeled by the $p$-th user. And $a_x^{(p)} = 1$ if pixel $x$ is labeled as salient and $a_x^{(p)} = 0$ otherwise. We define label consistency as the ratio between the number of pixels labeled as salient by all three people and the number of pixels labeled as salient by at least one of the people. It is formulated as
\begin{equation}
 C = \frac{\sum_x \left( \prod_{p=1}^{3}a_{x}^{(p)} \right)}{\sum_x \mathbf{1}\left( \sum_{p=1}^{3}a_{x}^{(p)} \neq 0 \right)}.
\end{equation}

We excluded those images with label consistency $C < 0.9$, and 4447 images remained. For each image that passed the label consistency test, we generated a ground truth saliency map from the annotations of three people. The pixelwise saliency label in the ground truth saliency map, $G = \{g_x|g_x \in \{0, 1\}\}$, is determined according to the majority label among the three people as follows,
\begin{equation}
g_x = \mathbf{1}\left(\sum_{p = 1}^{3}a_x^{(p)} \geq 2 \right).
\end{equation}

At the end, our new saliency dataset, called HKU-IS, contains 4447 images with high-quality pixelwise annotations. All the images in HKU-IS satisfy at least one of the above three criteria while 2888 (out of 5000) images in the MSRA dataset do not satisfy any of these criteria. In summary, 50.34\% images in HKU-IS have multiple disconnected salient objects while this number is only 6.24\% for the MSRA dataset; 21\% images in HKU-IS have salient objects touching the image boundary while this number is 13\% for the MSRA dataset; and the mean color contrast of HKU-IS is 0.69 while that of the MSRA dataset is 0.78.

\begin{figure*}[ht]
\begin{center}
   \includegraphics[width=1.0\textwidth]{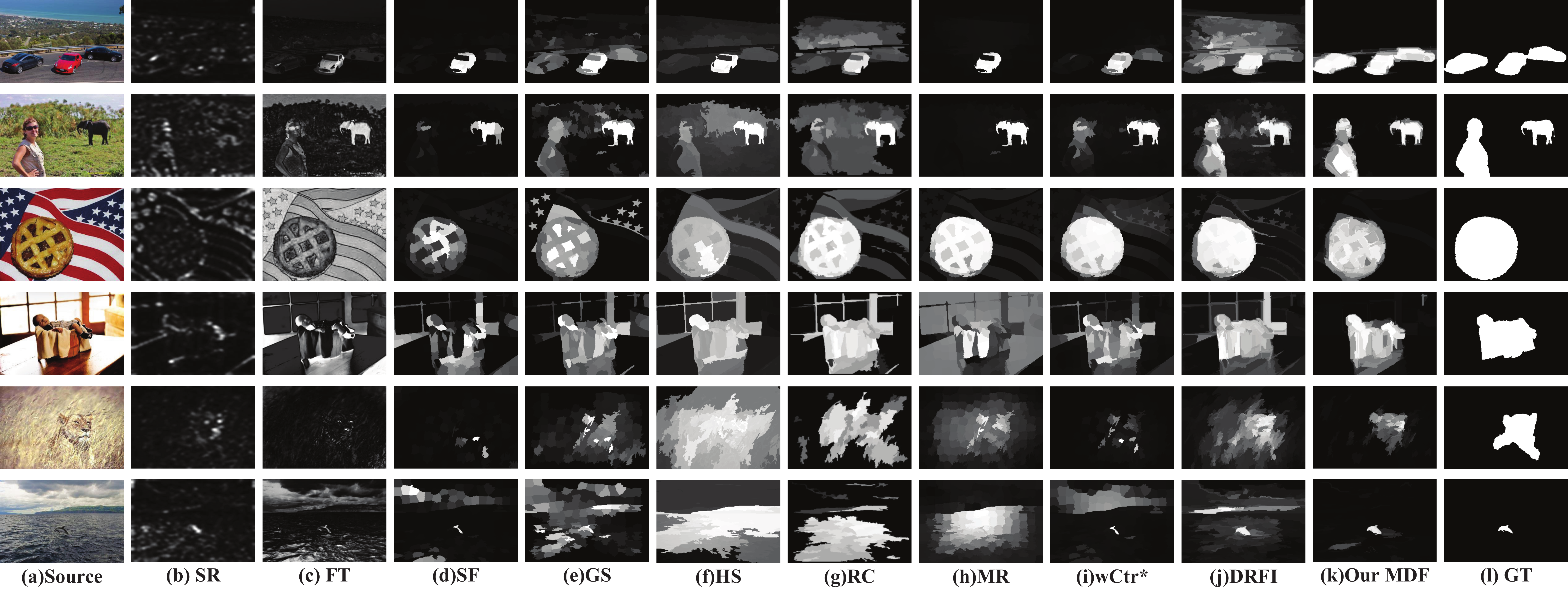}\vspace{-4mm}
\end{center}
   \caption{Visual comparison of saliency maps generated from 10 different methods, including ours (MDF). The ground truth (GT) is shown in the last column. MDF consistently produces saliency maps closest to the ground truth. We compare MDF against spectral residual (SR\cite{hou2007saliency}), frequency-tuned saliency (FT~\cite{achanta2009frequency}), saliency filters (SF~\cite{perazzi2012saliency}), geodesic saliency (GS~\cite{wei2012geodesic}), hierarchical saliency (HS~\cite{yan2013hierarchical}), regional based contrast (RC~\cite{ChengPAMI}), manifold ranking (MR~\cite{yang2013saliency}), optimized weighted contrast (wCtr$^*$~\cite{zhu2014saliency}) and  discriminative regional feature integration (DRFI~\cite{jiang2013salient}).}
\label{fig:long}
\end{figure*}

\section{Experimental Results}
\subsection{Dataset}
We have evaluated the performance of our method on several public visual saliency benchmarks as well as on our own dataset.

{\flushleft \textbf{MSRA-B}}\cite{liu2011learning}. This dataset has 5000 images, and is widely used for visual saliency estimation. Most of the images contain only one salient object. Pixelwise annotation was provided by \cite{jiang2013salient}.
\vspace{-0mm}
{\flushleft \textbf{SED}}\cite{alpert2007image}. It contains two subsets: SED1 and SED2. SED1 has 100 images each containing only one salient object while SED2 has 100 images each containing two salient objects.
\vspace{-0mm}
{\flushleft \textbf{SOD}}\cite{MartinFTM01}. This dataset has 300 images, and it was originally designed for image segmentation. Pixelwise annotation of salient objects in this dataset was generated by \cite{jiang2013salient}. This dataset is very challenging since many images contain multiple salient objects either with low contrast or overlapping with the image boundary.
\vspace{-0mm}
{\flushleft \textbf{iCoSeg}}\cite{batra2010icoseg}. This dataset was designed for co-segmentation. It contains 643 images with pixelwise annotation. Each image may contain one or multiple salient objects.
\vspace{-0mm}
{\flushleft \textbf{HKU-IS}}. Our new dataset contains 4447 images with pixelwise annotation of salient objects.

To facilitate a fair comparison with other methods, we divided the MSRA dataset into three parts as in \cite{jiang2013salient}, 2500 for training, 500 for validation and the remaining 2000 images for testing. Since other existing datasets are too small to train reliable models, we directly applied a trained model to generate their saliency maps as in \cite{jiang2013salient}.
We also divided HKU-IS into three parts, 2500 images for training, 500 images for validation and the remaining 1447 images for testing. The images for training and validation were randomly chosen from the entire dataset.

While it takes around 20 hours to train our deep neural network based prediction model for 15 image segmentation levels using the MSRA dataset, it only takes 8 seconds to detect salient objects in a testing image with 400x300 pixels on a PC with an NVIDIA GTX Titan Black GPU and a 3.4GHz Intel processor using our MATLAB code.

\begin{figure*}[t]
    \centerline{
	  \includegraphics[width = 0.236\textwidth]{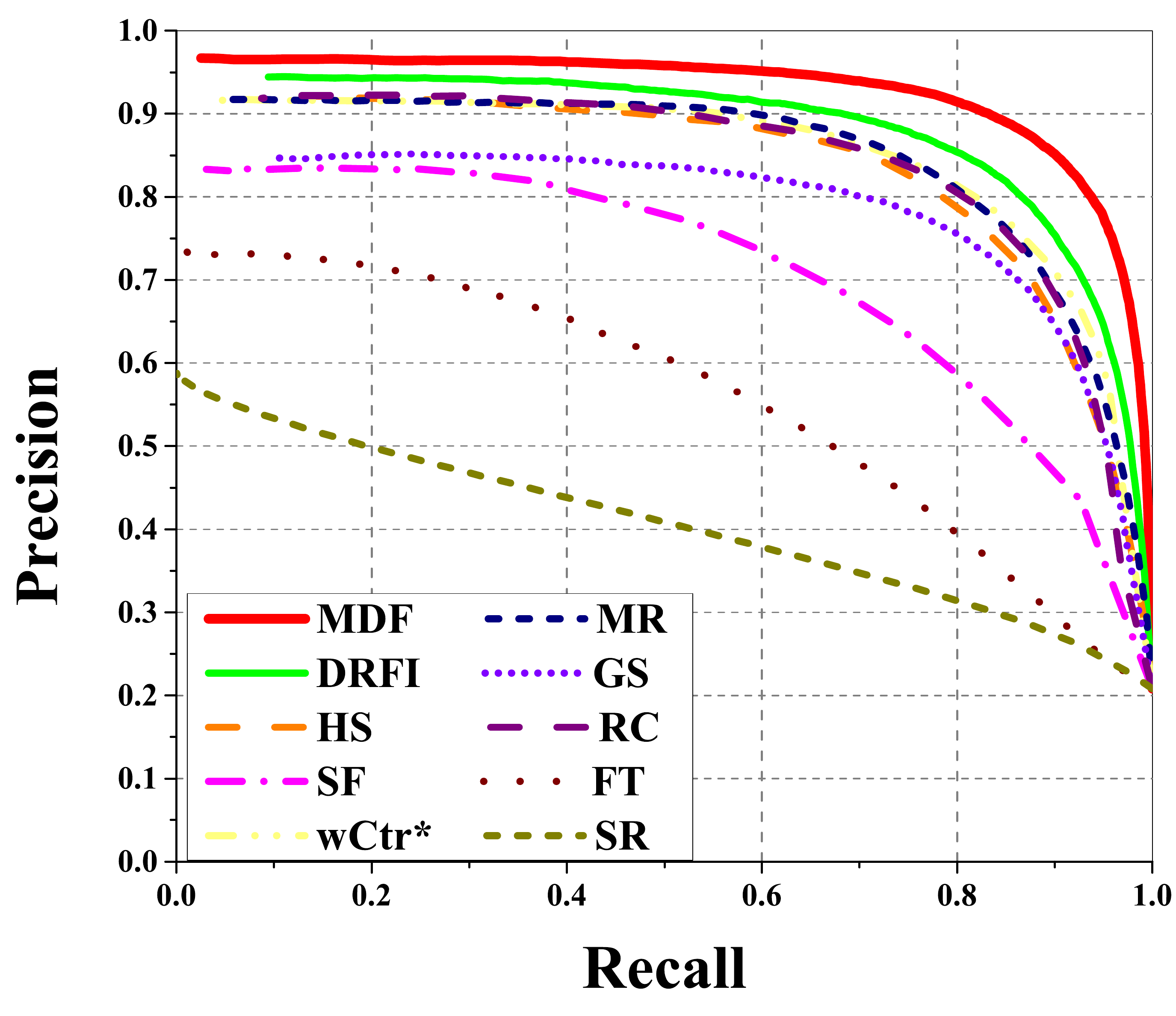}\hfill
	  \includegraphics[width = 0.236\textwidth]{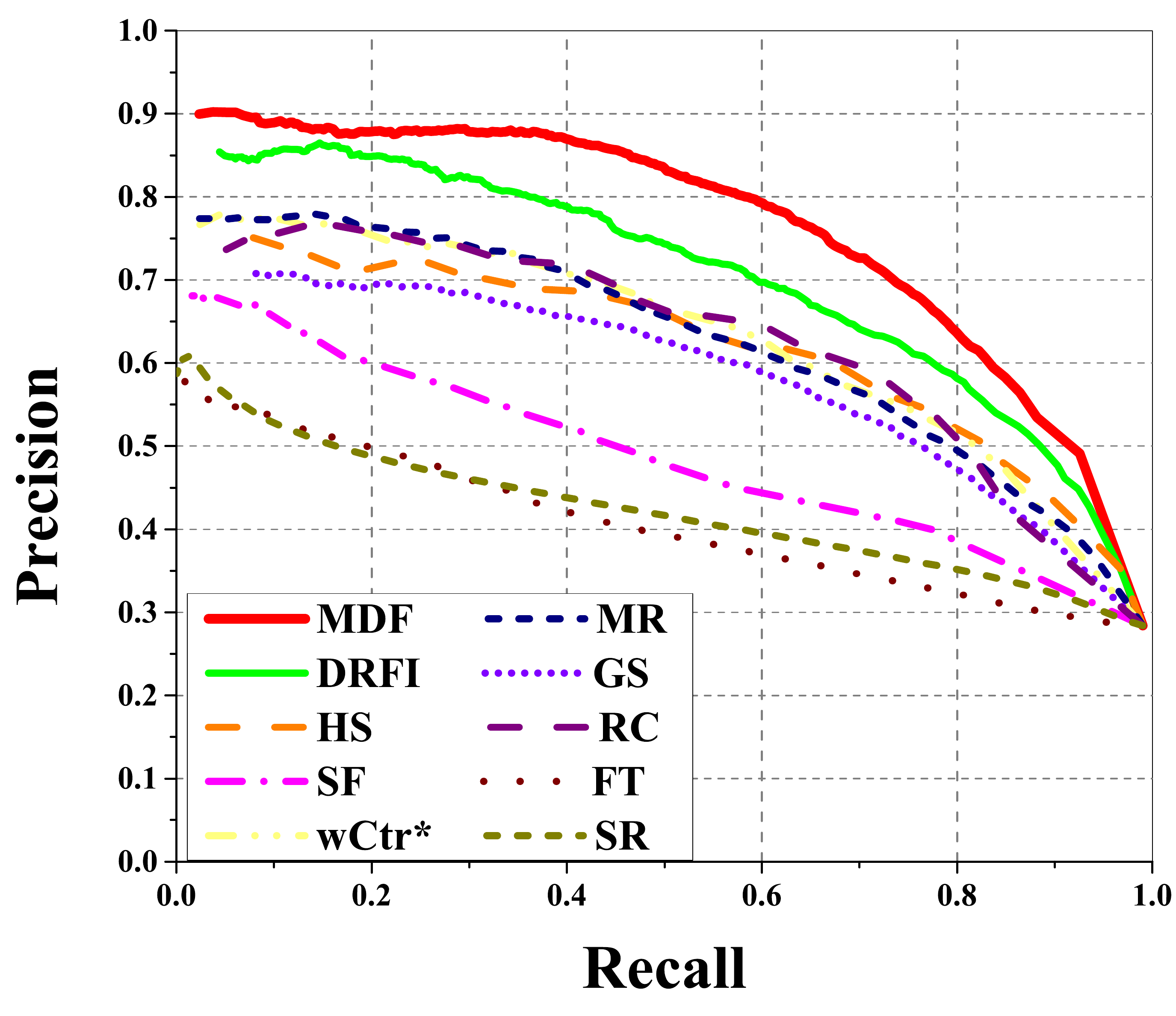}\hfill
	  \includegraphics[width = 0.236\textwidth]{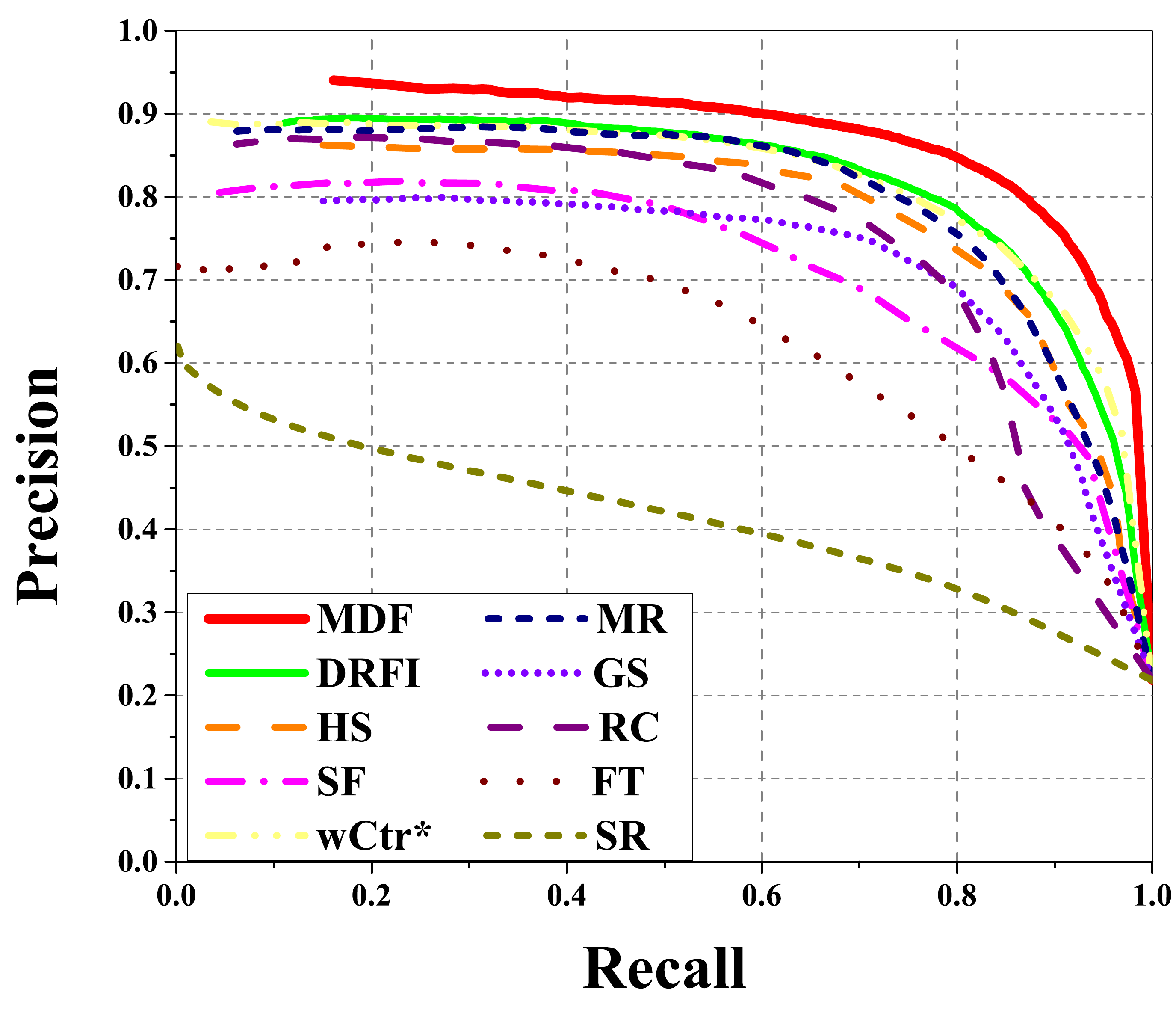}\hfill
	  \includegraphics[width = 0.236\textwidth]{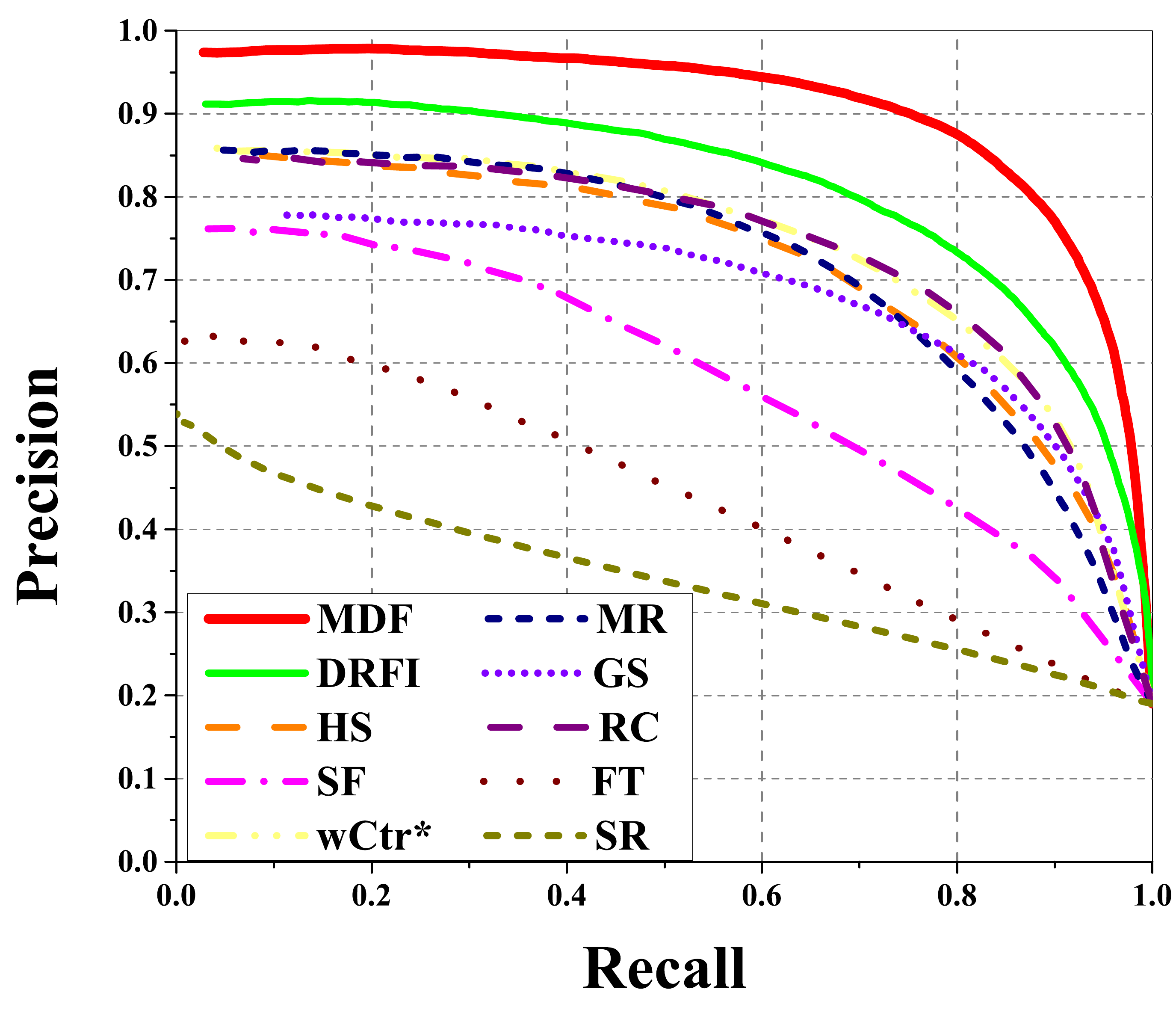}
	}
    \centerline{
	  \includegraphics[width = 0.236\textwidth]{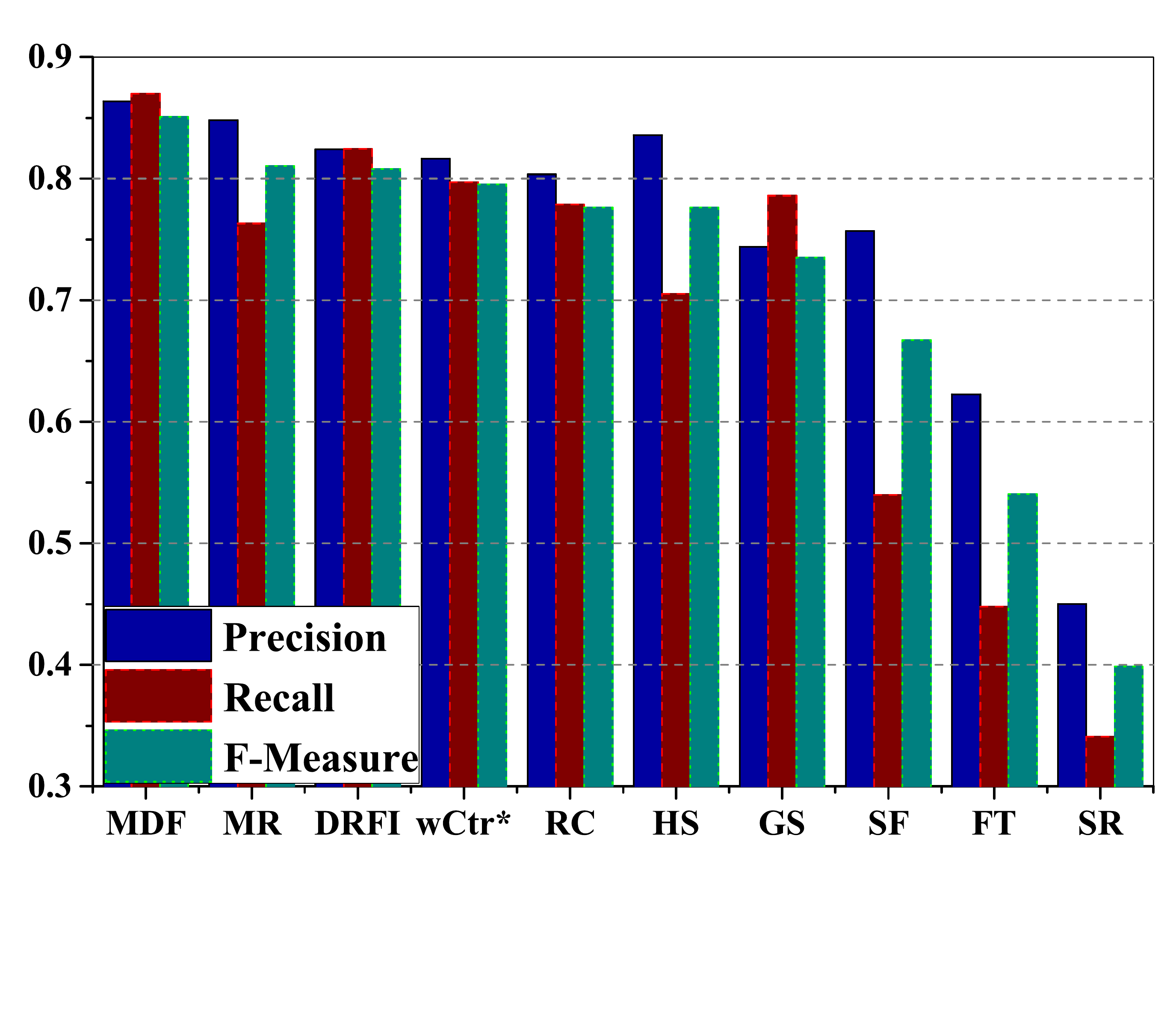}\hfill
	  \includegraphics[width = 0.236\textwidth]{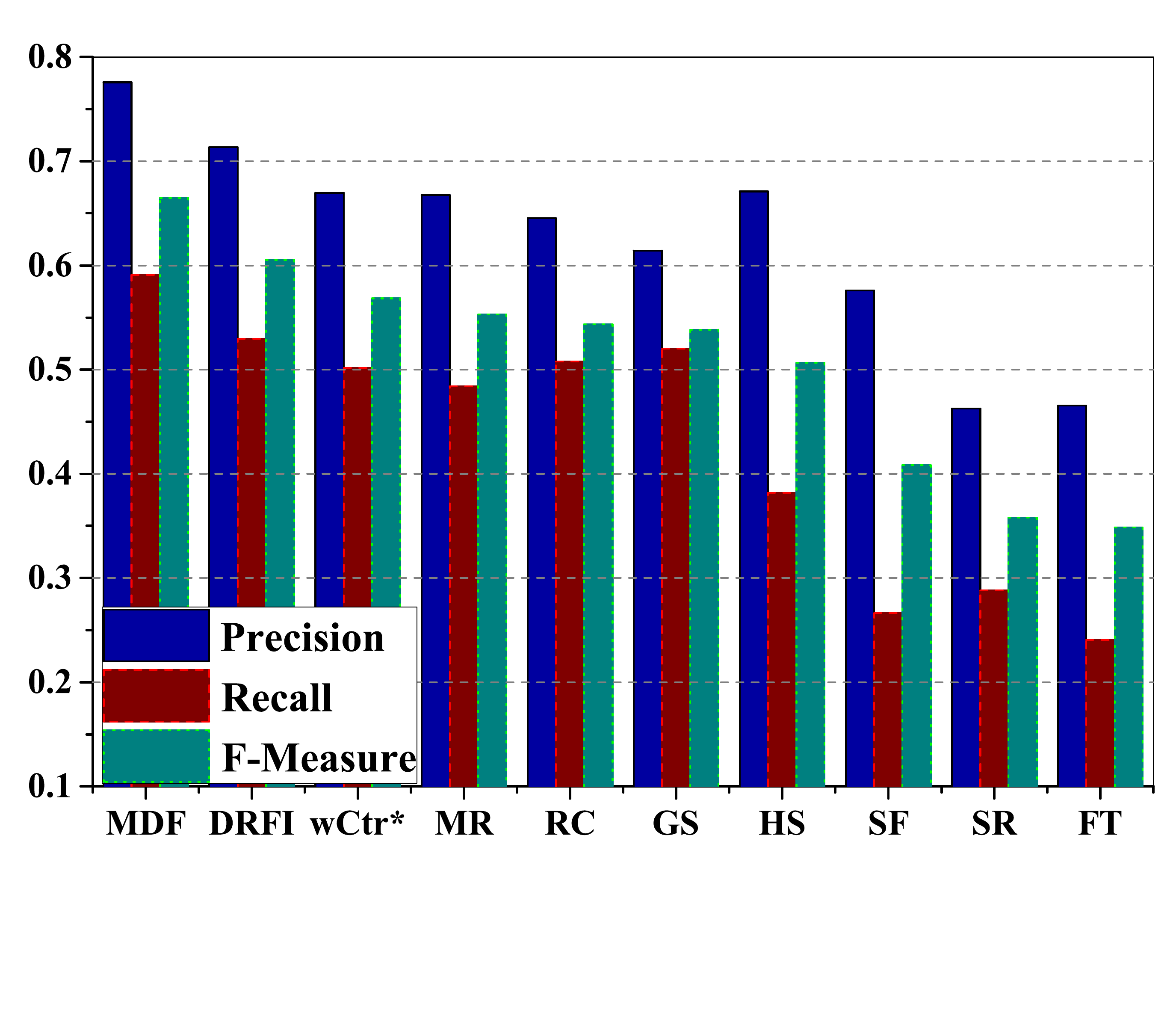}\hfill
	  \includegraphics[width = 0.236\textwidth]{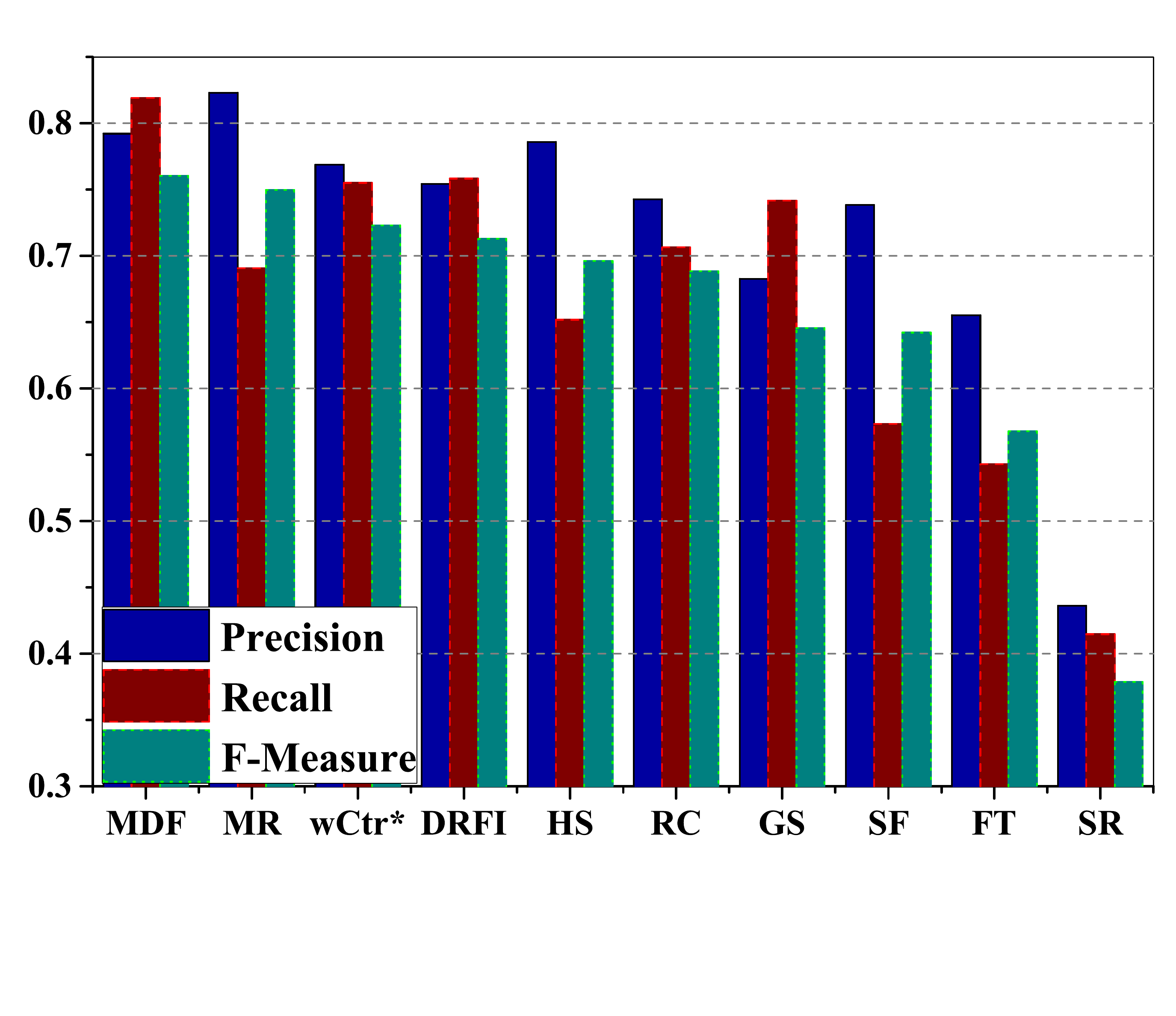}\hfill
	  \includegraphics[width = 0.236\textwidth]{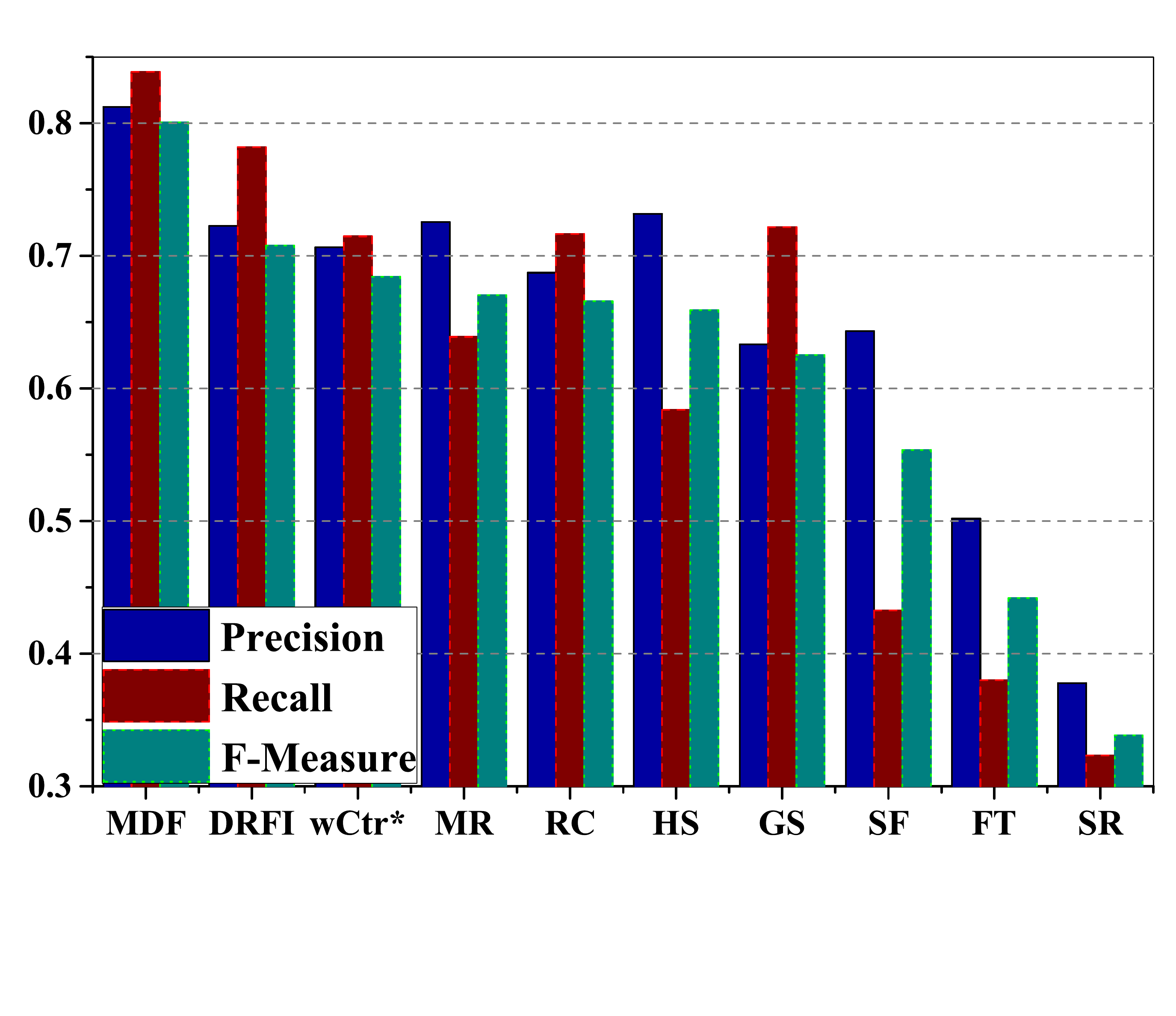}
	}
    \centerline{
	  \includegraphics[width = 0.236\textwidth]{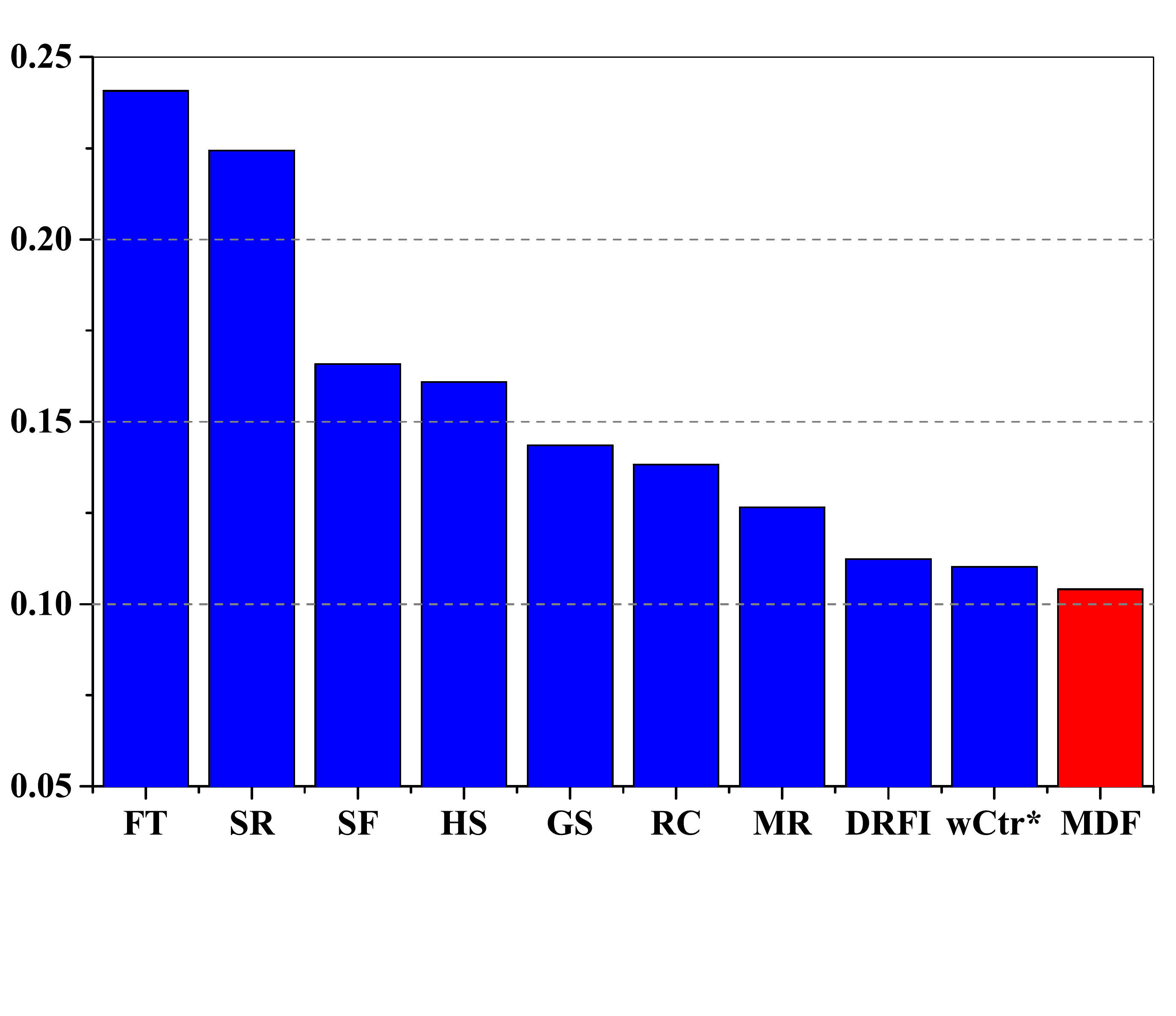}\hfill
	  \includegraphics[width = 0.236\textwidth]{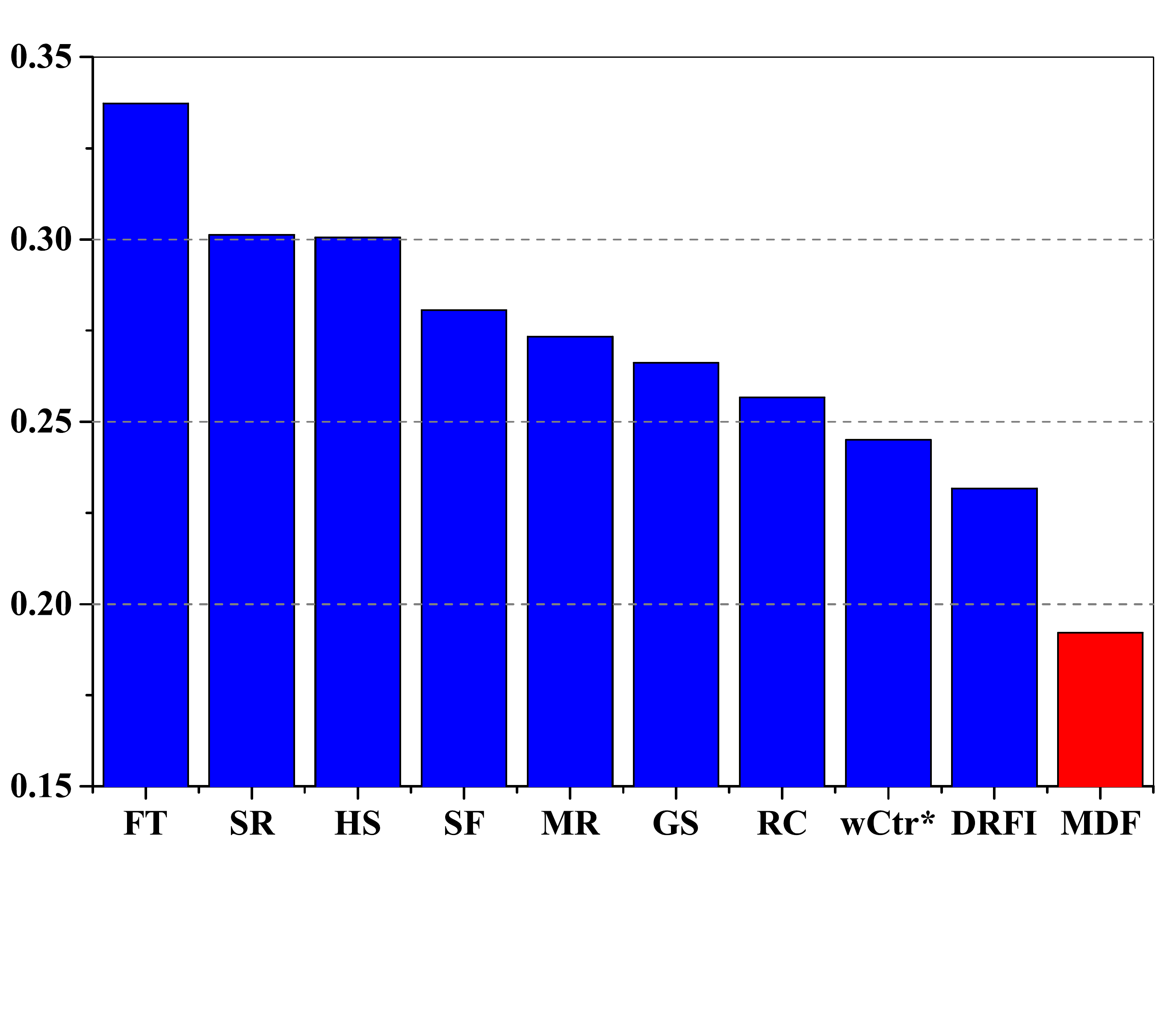}\hfill
	  \includegraphics[width = 0.236\textwidth]{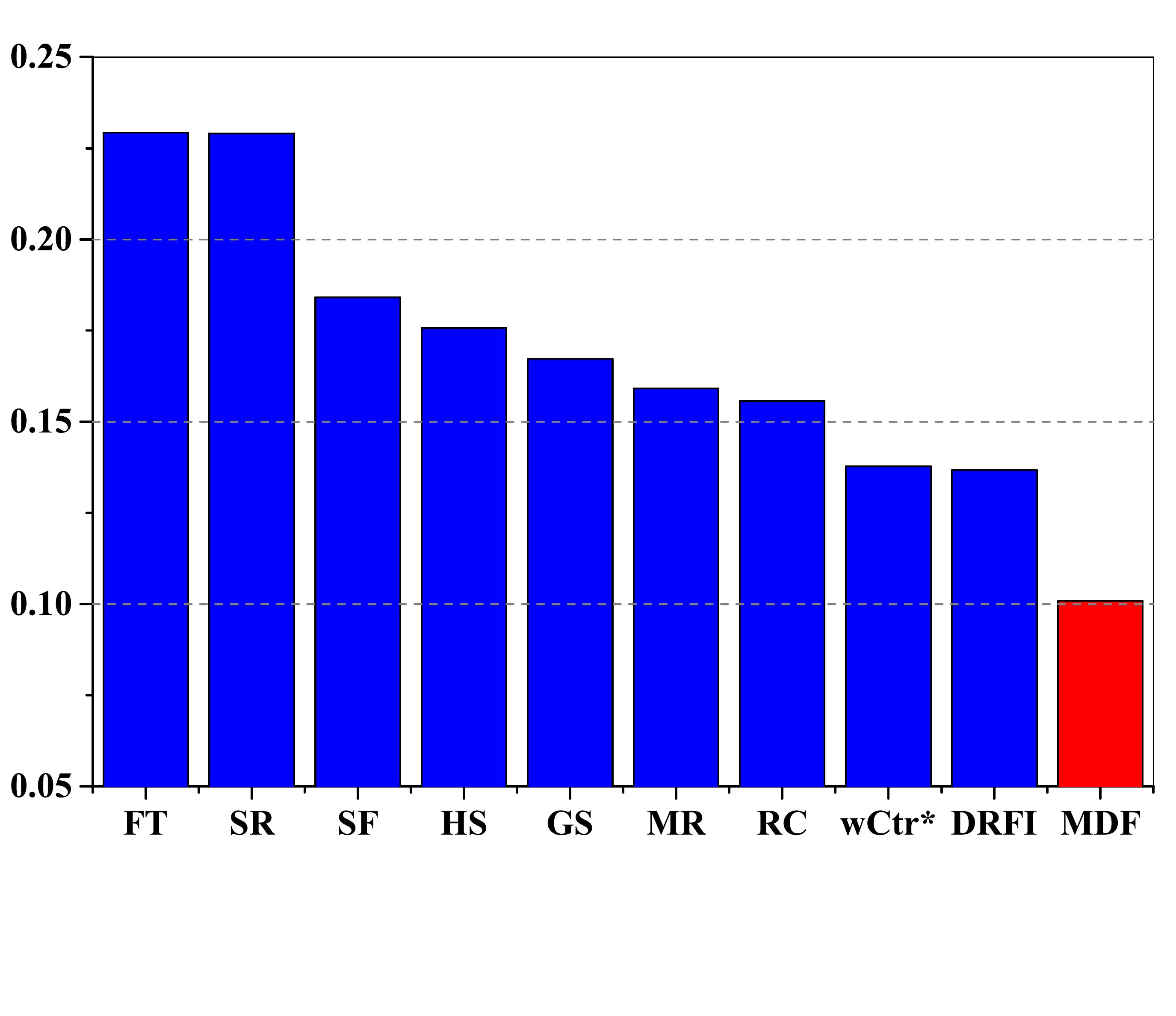}\hfill
	  \includegraphics[width = 0.236\textwidth]{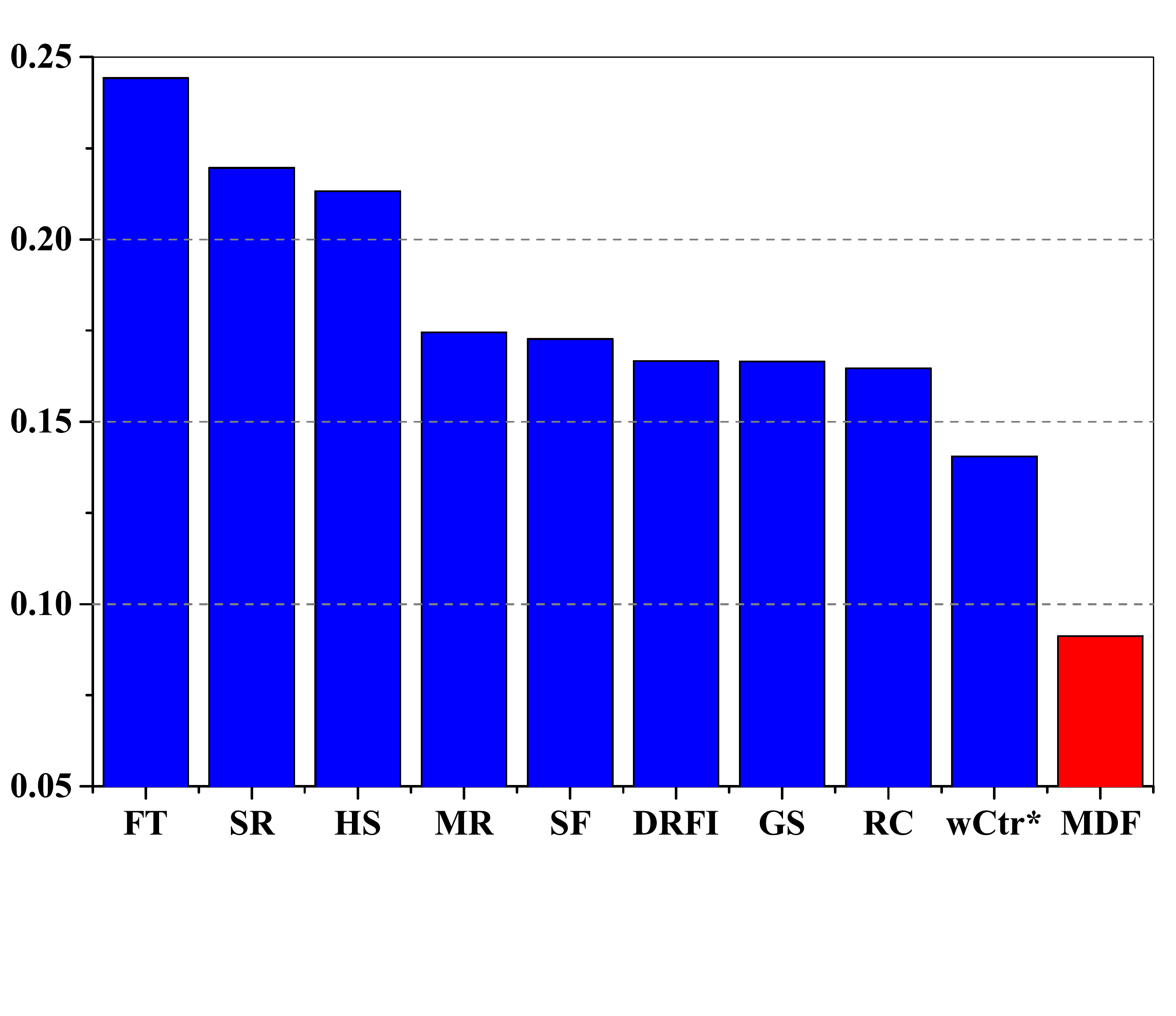}
	}\vspace{-4mm}
    \centerline{\hfill (a) \hfill\hfill (b) \hfill\hfill (c) \hfill\hfill (d) \hfill}\vspace{-1mm}
	\caption{Quantitative comparison of saliency maps generated from 10 different methods on 4 datasets. From left to right: (a) the MSRA-B dataset,  (b) the SOD dataset, (c) the iCoSeg dataset, and (d) our new HKU-IS dataset. From top to bottom: (1st row) the precision-recall curves of different methods, (2nd row) the precision, recall and F-measure using an adaptive threshold, and (3rd row) the mean absolute error.}\label{fig:comps}
\end{figure*}

\subsection{Evaluation Criteria}
Following \cite{achanta2009frequency,ChengPAMI}, we first use standard precision-recall curves to evaluate the performance of our method.
A continuous saliency map can be converted into a binary mask using a threshold, resulting in a pair of precision and recall values when the binary mask is compared against the ground truth. A precision-recall curve is then obtained by varying the threshold from $0$ to $1$. The curves are averaged over each dataset.

Second, since high precision and high recall are both desired in many applications, we compute the F-Measure\cite{achanta2009frequency} as
\begin{equation}
 F_{\beta} = \frac{(1+\beta^2)\cdot Precision \cdot Recall}{\beta^2\cdot Precision + Recall},
\end{equation}
where $\beta^2$ is set to 0.3 to weigh precision more than recall as suggested in \cite{achanta2009frequency}.
We report the performance when each saliency map is binarized with an image-dependent threshold proposed by \cite{achanta2009frequency}. This adaptive threshold is determined to be twice the mean saliency of the image:
\begin{equation}
 T_a = \frac{2}{W \times H} \sum_{x=1}^{W}\sum_{y=1}^{H}S(x,y),
\end{equation}
where $W$ and $H$ are the width and height of the saliency map $S$, and $S(x,y)$ is the saliency value of the pixel at $(x,y)$.
We report the average precision, recall and F-measure over each dataset.

Although commonly used, precision-recall curves have limited value because they fail to consider true negative pixels. For a more balanced comparison, we adopt the mean absolute error (MAE) as another evaluation criterion. It is defined as the average pixelwise absolute difference between the binary ground truth $G$ and the saliency map $S$~\cite{perazzi2012saliency},
\begin{equation}
MAE = \frac{1}{W\times H}\sum_{x=1}^{W}\sum_{y=1}^{H}|S(x,y) - G(x,y)|.
\end{equation}
MAE measures the numerical distance between the ground truth and the estimated saliency map, and is more meaningful in evaluating the applicability of a saliency model in a task such as object segmentation.

\begin{figure*}[ht]
    \centerline{
	  \includegraphics[width = 0.236\textwidth]{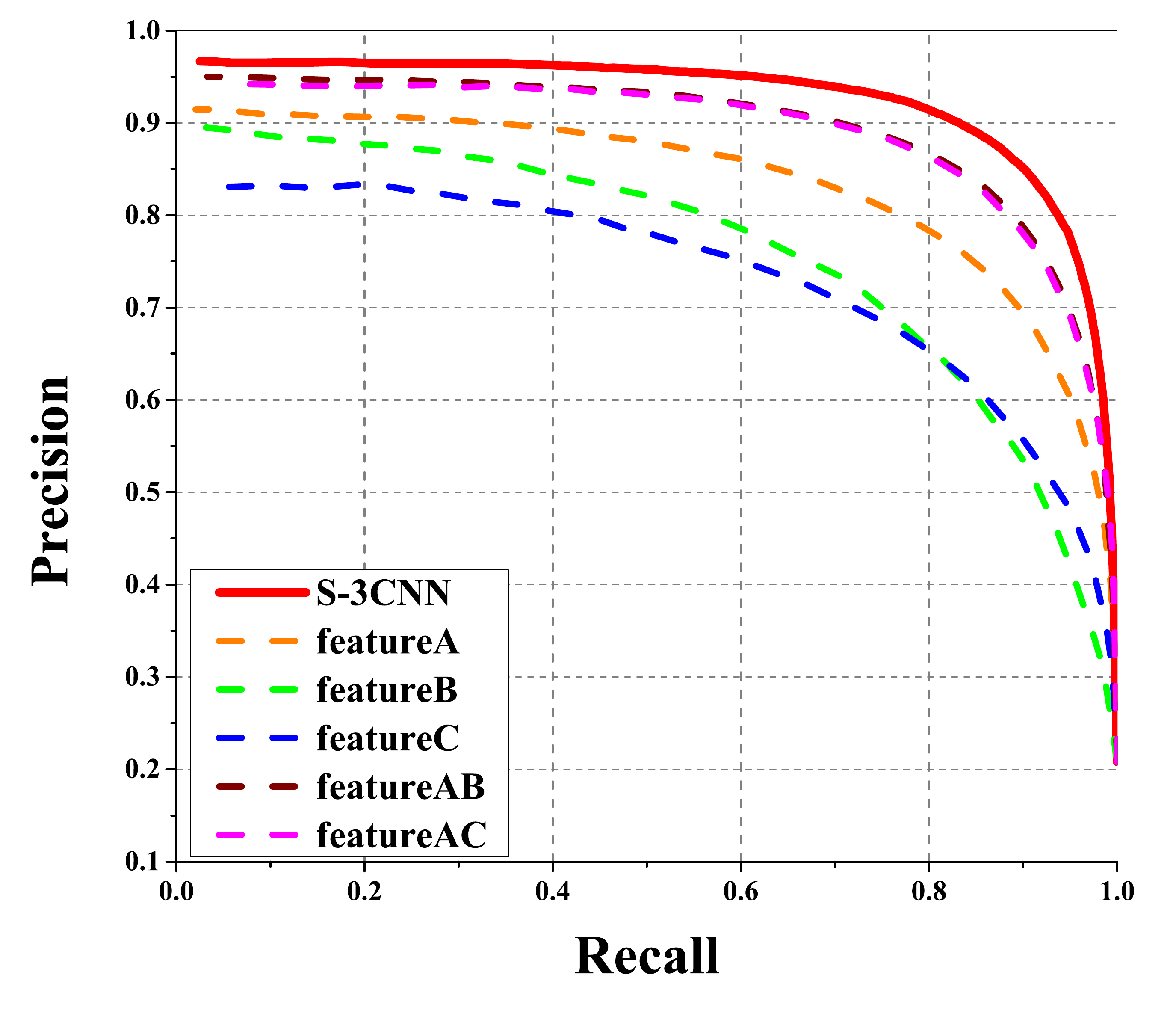}\hfill
	  \includegraphics[width = 0.236\textwidth]{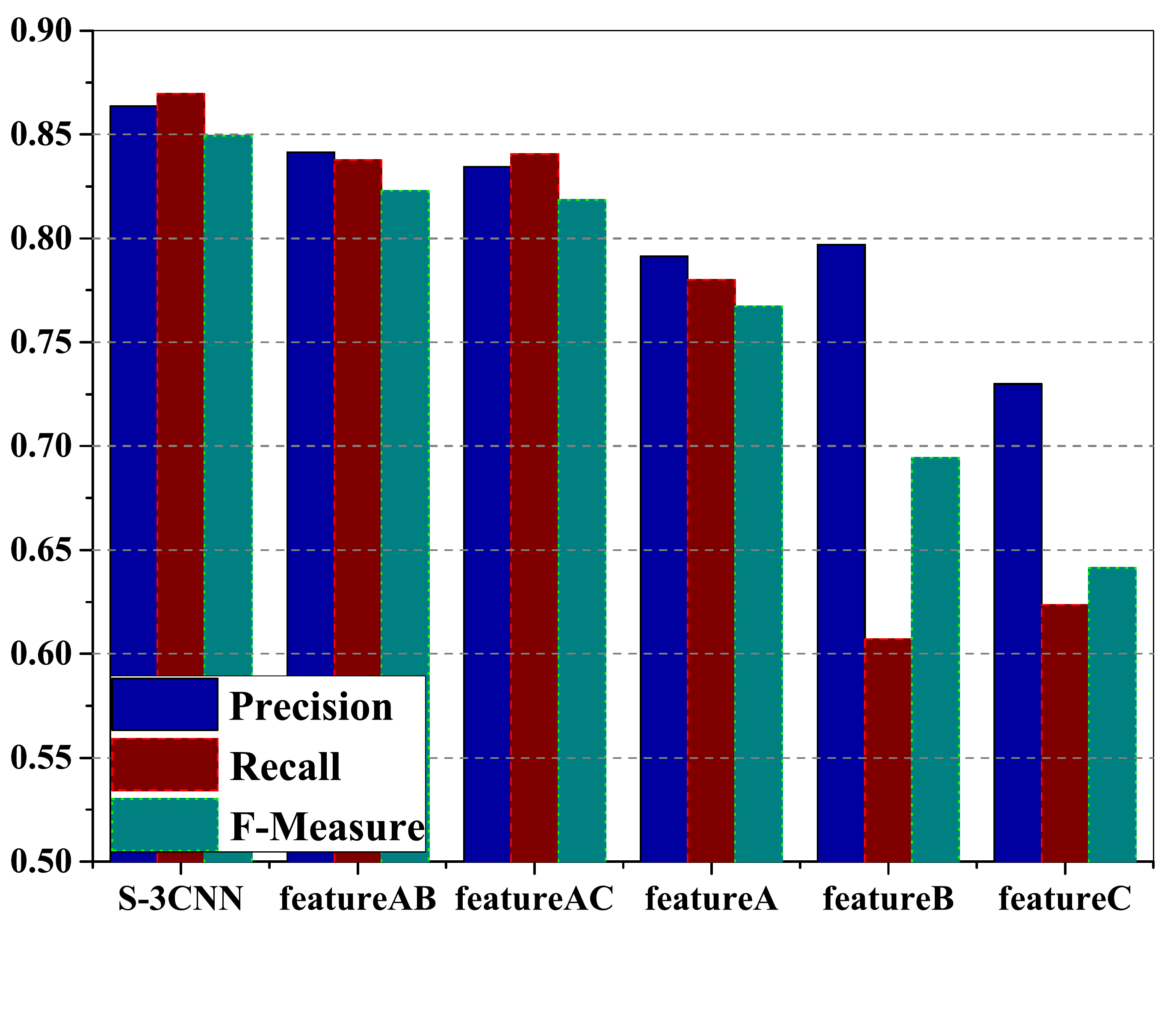}\hfill
	  \includegraphics[width = 0.236\textwidth]{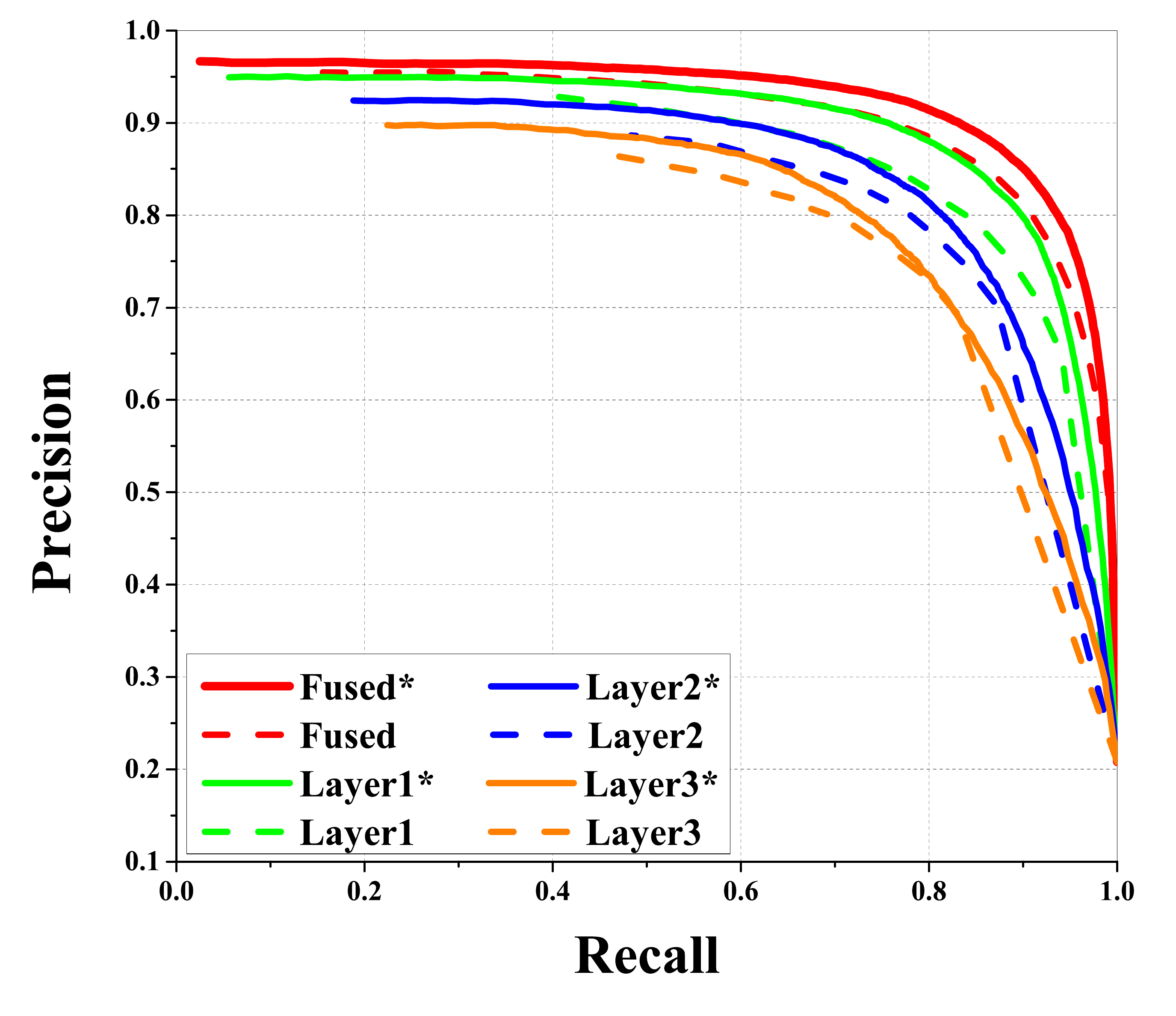}\hfill
	  \includegraphics[width = 0.236\textwidth]{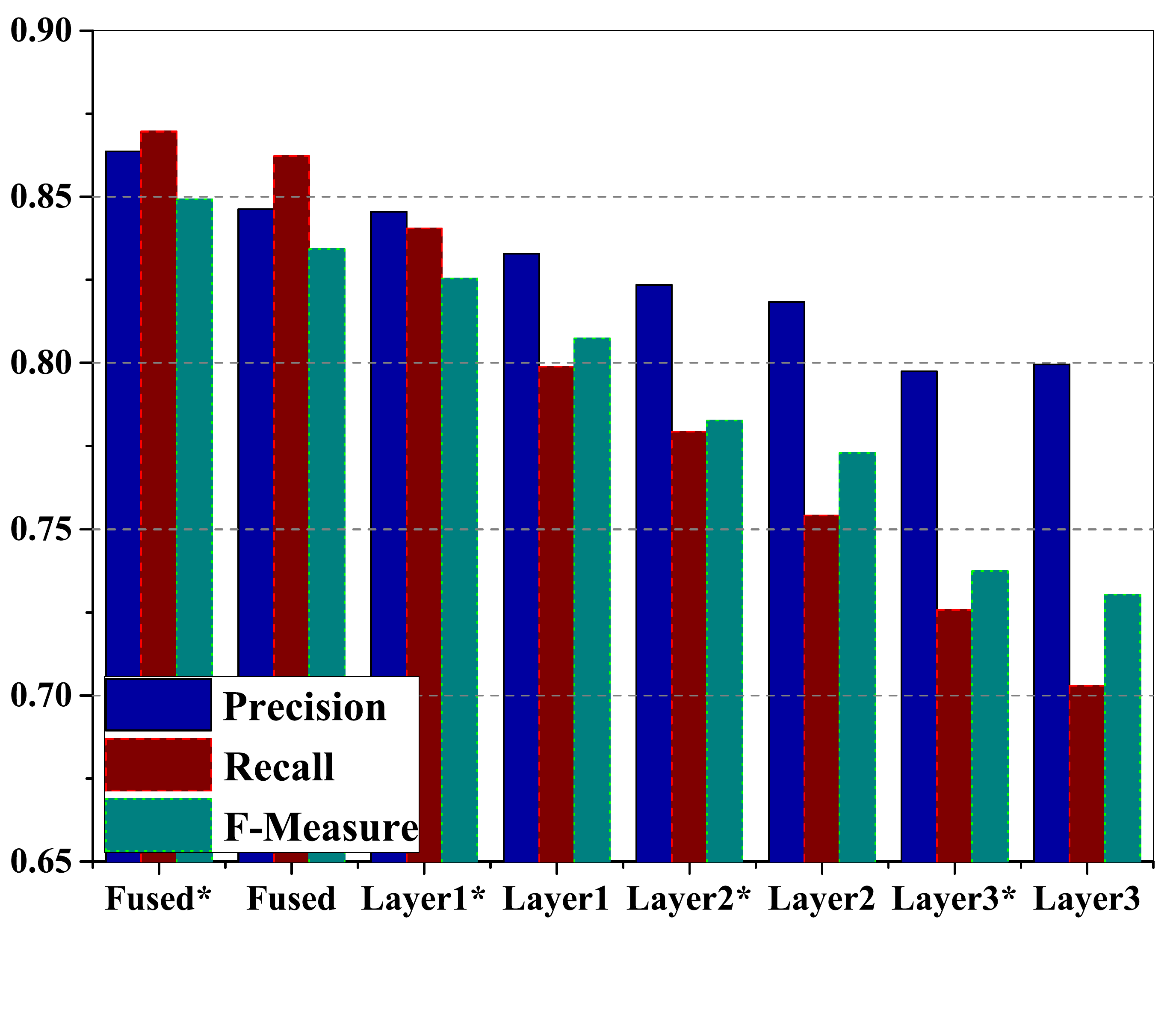}
	}\vspace{-0mm}
    \centerline{\hfill (a) \hfill\hfill (b) \hfill\hfill (c) \hfill\hfill (d) \hfill}\vspace{-1mm}
    \caption{Component-wise efficacy in our visual saliency model. (a) and (b) show the effectiveness of our S-3CNN feature. (a) shows the precision-recall curves of models trained on MSRA-B using different components of S-3CNN, while (b) shows the corresponding precision, recall and F-measure using an adaptive threshold. (c) and (d) show the effectiveness of spatial coherence and multilevel fusion. ``*'' refers to models with spatial coherence. ``Layer1", ``Layer2" and ``Layer3" refer to the three segmentation levels that have the highest single-level saliency prediction performance.}
\label{fig:analysis}
\end{figure*}

\subsection{Comparison with the State of the Art}
Let us compare our saliency model (MDF) with a number of existing state-of-the-art methods, including discriminative regional feature integration (DRFI)~\cite{jiang2013salient}, optimized weighted contrast (wCtr$^*$)~\cite{zhu2014saliency}, manifold ranking (MR)~\cite{yang2013saliency}, regional based contrast (RC)~\cite{ChengPAMI}, hierarchical saliency (HS)~\cite{yan2013hierarchical}, geodesic saliency (GS)~\cite{wei2012geodesic}, saliency filters (SF)~\cite{perazzi2012saliency}, frequency-tuned saliency (FT)~\cite{achanta2009frequency} and the spectral residual approach (SR)~\cite{hou2007saliency}. For RC, FT and SR, we use the implementation provided by \cite{ChengPAMI}; for other methods, we use original codes with recommended parameter settings.

A visual comparison is given in Fig. \ref{fig:long}. As can be seen, our method performs well in a variety of challenging cases, e.g., multiple disconnected salient objects (the first two rows), objects touching the image boundary (the second row), cluttered background (the third and fourth rows), and low contrast between object and background (the last two rows).

As part of the quantitative evaluation, we first evaluate our method using precision-recall curves. As shown in the first row of Fig. \ref{fig:comps}, our method achieves the highest precision in almost the entire recall range on all datasets. Precision, recall and F-measure results using the aforementioned adaptive threshold are shown in the second row of Figure \ref{fig:comps}, sorted by the F-measure. Our method also achieves the best performance on the overall F-measure as well as significant increases in both precision and recall. On the MSRA-B dataset, our method achieves 86.4\% precision and 87.0\% recall while the second best (MR) achieves 84.8\% precision and 76.3\% recall. Performance improvement becomes more obvious on HKU-IS. Compared with the second best (DRFI), our method increases the F-measure from 0.71 to 0.80, and achieves an increase of 9\% in precision while at the same time improving the recall by 5.7\%. Similar conclusions can also be made on other datasets. Note that the precision of certain methods, including MR\cite{yang2013saliency}, DRFI\cite{jiang2013salient}, HS\cite{yan2013hierarchical} and wCtr*\cite{zhu2014saliency}, is comparable to ours while their recalls are often much lower. Thus it is more likely for them to miss salient pixels. This is also reflected in the lower F-measure and higher MAE. Refer to the supplemental materials for the results on the SED dataset.

The third row of Fig. \ref{fig:comps} shows that our method also significantly outperforms other existing methods in terms of the MAE measure, which provides a better estimation of the visual distance between the predicted saliency map and the ground truth. Our method successfully lowers the MAE by 5.7\% with respect to the second best algorithm (wCtr*) on the MSRA-B dataset. On two other datasets, iCoSeg and SOD, our method lowers the MAE by 26.3\% and 17.1\% respectively with respect to the second best algorithms. On HKU-IS, which contains more challenging images, our method significantly lowers the MAE by 35.1\% with respect to the second best performer on this dataset (wCtr*).

In summary, the improvement our method achieves over the state of the art is substantial. Furthermore, the more challenging the dataset, the more obvious the advantages because our multiscale CNN features are capable of characterizing the contrast relationship among different parts of an image.

\subsection{Component-wise Efficacy}
\paragraph{Effectiveness of S-3CNN} As discussed in Section~\ref{sec:feature}, our multiscale CNN feature vector, S-3CNN, consists of three components, A, B and C. To show the effectiveness and necessity of these three parts, we have trained five additional models for comparison, which respectively take $feature$ A only, $feature$ B only, $feature$ C only, concatenated A and B, and concatenated A and C. These five models were trained on MSRA-B using the same setting as the one taking S-3CNN. Quantitative results were obtained on the testing images in the MSRA-B dataset. As shown in Fig. \ref{fig:analysis}, the model trained using S-3CNN consistently achieves the best performance on average precision, recall and F-measure. Models trained using two components perform much better than those trained using a single component.
These results demonstrate that the three components of our multiscale CNN feature vector are complementary to each other, and the training stage of our saliency model is capable of discovering and understanding region contrast information hidden in our multiscale features.

\paragraph{Spatial Coherence} In Section~\ref{sec:coherence}, spatial coherence was incorporated to refine the saliency scores from our CNN-based model. To validate its effectiveness, we have evaluated the performance of our final saliency model with and without spatial coherence using the testing images in the MSRA-B dataset. We further chose the three segmentation levels that have the highest single-level saliency prediction performance, and compared their performance with spatial coherence turned on and off. The resulting precision-recall curves are shown in Fig. \ref{fig:analysis}. It is evident that spatial coherence clearly improves the accuracy of our models.

\paragraph{Multilevel Decomposition} Our method exploits information from multiple levels of image segmentation. As shown in Fig. \ref{fig:analysis}, the performance of a single segmentation level is not comparable to the performance of the fused model. The aggregated saliency map from 15 levels of image segmentation improves the average precision by $2.15\%$ and at the same time improves the recall rate by $3.47\%$ when it is compared with the result from the best-performing single level.


\section*{Acknowledgments}
We would like to thank Sai Bi, Wei Zhang, and Feida Zhu for their help during the construction of our dataset. The first author is supported by Hong Kong Postgraduate Fellowship.

{\small
\bibliographystyle{ieee}
\bibliography{ref,saliency}
}

\end{document}